\documentclass[runningheads]{llncs}
\usepackage{graphicx}
\usepackage{algorithm}
\usepackage{algpseudocode}
\usepackage{caption}
\usepackage{subcaption}
\usepackage{amsmath}
\usepackage{amsfonts}
\usepackage{epstopdf}
\begin{document}
\title{Self-Supervised Anomaly Detection of Rogue Soil Moisture Sensors}
\author{Boje Deforce\inst{1}\orcidID{0000-0001-6530-9752} \and
Bart Baesens\inst{1,2}\orcidID{0000-0002-5831-5668} \and
Jan Diels\inst{3}\orcidID{0000-0002-0317-8280} \and
Estefanía {Serral Asensio}\inst{1}\orcidID{0000-0001-7579-910X}}
\authorrunning{B. Deforce et al.}
\institute{Research Center for Information Systems Engineering (LIRIS), KU Leuven, Belgium
\email{\{boje.deforce, bart.baesens, estefania.serralasensio\}@kuleuven.be} \and 
Department of Decision Analytics and Risk, University of Southampton, United Kingdom \and Division of Soil and Water Management, KU Leuven, Belgium \email{jan.diels@kuleuven.be}}
\maketitle              % typeset the header of the contribution
\begin{abstract}
IoT data is a central element in the successful digital transformation of agriculture. However, IoT data comes with its own set of challenges. E.g., the risk of data contamination due to rogue sensors. A sensor is considered rogue when it provides incorrect measurements over time. To ensure correct analytical results, an essential preprocessing step when working with IoT data is the detection of such rogue sensors. Existing methods assume that well-behaving sensors are known or that a large majority of the sensors is well-behaving. However, real-world data is often completely unlabeled and voluminous, calling for self-supervised methods that can detect rogue sensors without prior information. We present a self-supervised anomalous sensor detector based on a neural network with a contrastive loss, followed by DBSCAN. A core contribution of our paper is the use of Dynamic Time Warping in the negative sampling for the triplet loss. This novelty makes the use of triplet networks feasible for anomalous sensor detection. Our method shows promising results on a challenging dataset of soil moisture sensors deployed in multiple pear orchards.

\keywords{Contrastive learning \and Anomaly detection \and IoT data.}
\end{abstract}
\section{Introduction}
 IoT sensors are central to the successful digital transformation of agriculture~\cite{10.3390/s21113758}. However, IoT sensors used in an agricultural context are often directly exposed to harsh conditions that can make sensor corruption more likely and unpredictable, even more so when dealing with battery-powered sensors~\cite{Zhang2010OutlierSurvey}. Erroneous sensor data can ultimately cause a cascaded effect in automated decision systems such as e.g. smart irrigation~\cite{Barreto_Amaral_2018}. As such, it is essential to detect rogue sensors to ensure data quality. Rogue sensors are sensors that show anomalous behavior. Formally, rogue sensors can be considered as sensors which show ``measurable consequences of an unexpected change in state of a system which is outside of its local or global norm"~\cite{Cook2020AnomalySurvey}. The sheer volume of data produced by IoT sensors on a daily basis has made manual inspection impossible, resulting in a growing emphasis in the literature on efficient analytical methods to deal with IoT data~\cite{Cook2020AnomalySurvey,Mohammadi2018DeepSurvey}.To this end, we propose a fully self-supervised anomalous sensor detector based on a triplet network with a novel negative sampling method relying on dynamic time warping (DTW), followed by Density-based spatial clustering of applications with noise (DBSCAN).
\begin{figure}[!h]
\includegraphics[width=\linewidth]{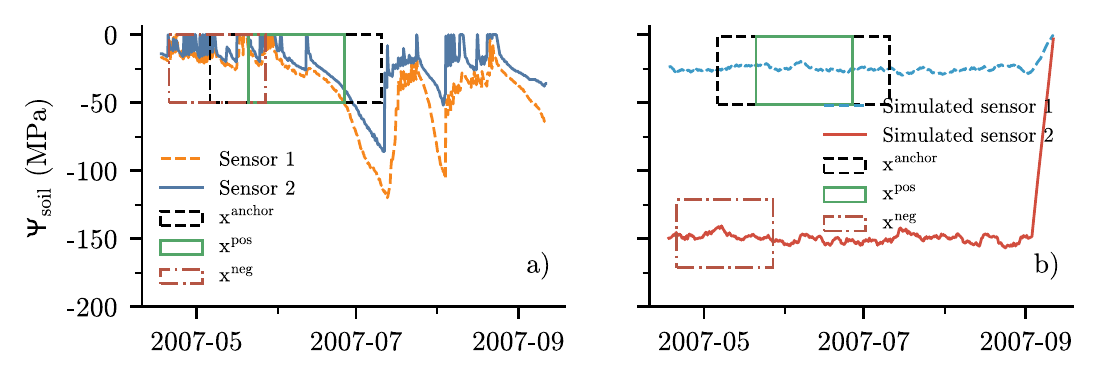}
\caption{b) shows a desirable (simulated) situation where $x^{neg}$ differs substantially - and will be separated in latent space - from $x^{anchor}$ and $x^{pos}$ as opposed to a), our real-world data, where $x^{anchor}$, $x^{pos}$, and $x^{neg}$ show strong similarity. (All samples are selected following the negative sampling approach in~\cite{Franceschi2019UnsupervisedSeries})} \label{fig1}
\end{figure}
\section{Related work}
%
%While there is a general lack of existing work on ML-driven anomaly detection in smart agriculture \cite{Abdallah_Lee_Raghunathan_Mousoulis_Sutherland_Bagchi_2021,Barreto_Amaral_2018}, 
In recent work two types of relevant anomaly detection in agriculture emerged. 
One type focuses on specific agricultural anomalies. E.g. in~\cite{Mouret_Albughdadi_Duthoit_Kouam_Rieu_Tourneret_2021}, satellite imagery is used to detect crop anomalies using an isolation forest with a pre-defined contamination rate. In~\cite{Moso_Cormier_Runz_Fouchal_Wandeto_2021}, an unsupervised anomaly detector is proposed which also includes an isolation forest a.o. techniques to detect anomalies in data streams from trajectories of combine harvesters and to detect crop anomalies. DeepAnomaly was introduced in~\cite{Christiansen_Nielsen_Steen_Jorgensen_Karstoft_2016} and uses a combo of a background substraction algorithm and a convolutional neural network (CNN) for real-time detection of obstacles and anomalies in agricultural fields for e.g. autonomous tractors. The other type focuses on the broader issue of data quality in sensor networks deployed in an agricultural context.
%Next, we describe some representative examples. 
%While the aforementioned works tackle specific agricultural anomalies, few focus on the broader issue of data quality in IoT-networks deployed in an agricultural context~\cite{Abdallah_Lee_Raghunathan_Mousoulis_Sutherland_Bagchi_2021}. 
To this end,~\cite{Abdallah_Lee_Raghunathan_Mousoulis_Sutherland_Bagchi_2021} focuses on detecting anomalies in IoT-data by comparing predicted values of a given sensor with the observed values. Based on a set threshold, exceeding values are then tagged as anomalous. They use an Autoregressive Integrated Moving Average (ARIMA) model and Long Short-Term memory (LSTM) neural network with the transfer learning. A limitation of this approach is that the majority of the sensors need to be non-anomalous in order to train their models.
%for their model to learn what non-anomalous data looks like in order to forecast it correctly.
~\cite{Ou_Chen_Huang_Huang_2020} uses linear regression to obtain a trend-line in time-windows of various sensor data (e.g. humidity, soil temperature, ...). Subsequently, using quartile thresholds, they define a given time-window as anomalous or not. Here too, it is implicitly assumed that the majority of sensors in a given window is non-anomalous. However,~\cite{Moso_Cormier_Runz_Fouchal_Wandeto_2021} highlight how real-world scenarios typically lack information on which sensors are anomalous and which are not, emphasising the importance of having fully self-supervised anomaly detectors. Vilenski et al.~\cite{Vilenski_Bak_Rosenblatt_2019} come perhaps closest to our approach where they built a novel unsupervised anomaly detection pipeline for detecting data quality issues in dendrometer sensor networks that is also transferable to other use-cases. However, their method requires a lot of expertise and effort into preprocessing the data. Moreover, it is unclear whether this method is capable of dealing with misaligned time-series which can occur due to e.g. field variability. Hence, there is a need for fully self-supervised anomaly detectors that can deal with raw (aligned or misaligned) sensory data.
\section{Anomaly detection with triplet dilated CNN and DBSCAN}\label{sec:triploss}
We introduce a contrastive self-supervised anomalous sensor detector with a triplet loss based on DTW which can detect anomalous sensors without any prior assumptions about the distinction of anomalous vs. non-anomalous data or the level of data contamination. Triplet networks have been successful in computer vision \cite{Chechik2010LargeRanking,Schroff_Kalenichenko_Philbin_2015} and natural language processing~\cite{Mikolov2013DistributedCompositionality} but were only recently introduced for time-series clustering and classification~\cite{Franceschi2019UnsupervisedSeries}. 
A triplet network allows to learn explicit discriminative embeddings $\mathbf{Z}$ of a given input $\mathbf{Y}\in \mathbb{R}^{n \times d}$. These embeddings typically live in a lower-dimensional space such that 
$\mathbf{Z}\in \mathbb{R}^{n \times p}$ with $p < d$. 
Formally, the triplet network $f$ is a mapping from the real data space to the latent space $f: \mathbf{Y} \to \mathbf{Z}$. The objective function of a triplet network embodies a triplet loss. A triplet loss consists of three main components: an anchor $x_i^{anchor}$, a positive sample $x_i^{pos}$, and a negative sample $x_i^{neg}$ (see also Eq.(\ref{eq1}) and Figure \ref{fig1}). The triplet loss will force the Euclidean distance between $x_i ^{anchor}$ and $x_i ^{pos}$ to be small while it will force $x_i ^{neg}$ to be far away from the anchor and the positive sample. Previous work for contrastive learning in time-series~\cite{Franceschi2019UnsupervisedSeries} assumes sufficient variety in the data such that $x_i ^{neg}$ will be substantially different from $x_i ^{anchor}$ and $x_i ^{pos}$ by random sampling (similar to word2vec~\cite{Mikolov2013DistributedCompositionality}). However, Figure \ref{fig1} demonstrates how this can break down when many time-series are similar such as in agricultural fields where many sensors measure the same concept. In this case, sampling at random can result in a situation where pushing $x_i ^{neg}$ far away from $x_i ^{anchor}$ and $x_i ^{pos}$ is illogical.

We propose instead a novel triplet loss sampling technique based on DTW-distance~\cite{Sakoe1978DynamicRecognition,Giorgino2009ComputingPackage}. DTW is an algorithm that finds the optimal alignment between 2 temporal unaligned sequences. The resulting DTW-distance is the sum of distances between the aligned elements, typically expressed as the Euclidean distance~\cite{Giorgino2009ComputingPackage}. We select $x_i ^{anchor}$, $x_i ^{pos}$ from a given time-series $y_i$ such that $x_i ^{pos}$ is a subsequence of $x_i ^{anchor}$ and such that $|x_i ^{anchor}| \leq |y_i|$. However, to select $x_i ^{neg}$ we measure the DTW-distance between $y_i$ (from which the anchor and positive were selected) and all other available time-series. We then proceed with the $K$ furthest neighbors of $y_i$ (in terms of DTW-distance) to select the negative samples. Formally, this results in algorithm \ref{alg:sampling} (for one epoch), inspired by~\cite{Franceschi2019UnsupervisedSeries}.
\begin{algorithm}[t]
\caption{Choosing $x^{anchor}$, $x^{pos}$, and $K$ negative samples $x_k^{neg}$}\label{alg:sampling}
\begin{algorithmic}[1]
\For{$i \in [1,N] \textbf{ with }s_i=\textrm{size}(y_i)$}
    \State pick $s^{pos}=\textrm{size}(x^{pos})$ in $[  1, s_i ]$ and $s^{anchor}=\textrm{size}(x^{anchor})$ in $[ s^{pos}, s_i ]$ uniformly at random;
    
    \State pick $x^{anchor}$ uniformly at random among subsequences of $y_i$ of length $s^{anchor}$;
    
    \State pick $x^{pos}$ uniformly at random among subsequences of $x^{anchor}$ of length $s^{pos}$;
    %new part
    \State compute DTW-distance $DTWDistance(y_i, y_j)$ for $j = 1, 2, ..., N$ and $j\ne i$;
    \State select top $K$ instances ordered by DTW-distance in descending order;
    \State pick $s_k^{neg}=\textrm{size}(x_k^{neg})$ in $[ 1, \textrm{size}(y_k) ]$ uniformly at random;
    \State pick $x_k^{neg}$ among subsequences of $y_k$ of length $s_k^{neg}$ for $k \in [ 1, K ]$;
    
\EndFor
\end{algorithmic}
\end{algorithm}
As such, we ensure that the negative samples are selected from time-series that are far away (by DTW-distance) from the time-series where the anchor and the positive sample were selected from, even when time-series are misaligned. We use these as input to the triplet network with the following objective loss function~\cite{Franceschi2019UnsupervisedSeries,Mikolov2013DistributedCompositionality}: 
\begin{equation}\label{eq1}
\resizebox{.93\hsize}{!}{$- log \left( \sigma \left(f\left(x^{anchor}, \theta \right)^\top f \left(x^{pos}, \theta \right)  \right) \right)
     - \sum_{k=1}^{K} log \left(\sigma \left(- f \left( x^{anchor}, \theta \right)^\top f \left(x_k^{neg}, \theta \right) \right)\right)$}
\end{equation}

%Our novel triplet loss sampling technique allows to learn discriminative embeddings that can be used effectively to distinguish between anomalous and non-anomalous time-series, even when many time-series are similar (see also Figure \ref{fig1}a).
For the network-architecture, we rely on a dilated CNN~\cite{Bai2018AnModeling} (see also section \ref{sec:dilated_cnn}). The output of the dilated CNN is then fed to a global max pooling layer and a fully connected layer to result in a vector of a fixed pre-determined size. The triplet network yields learned representations for each sensor's time-series. While these learned features are already discriminative by definition, a final anomaly detection step is required to objectively detect which rogue sensors/groups of rogue sensors emerge. To achieve this, we apply DBSCAN on the learned embeddings (see section \ref{sec:anomaly_detection}).
\subsection{Anomaly detection on learned representations}\label{sec:anomaly_detection}
\begin{figure}[h]
    \centering
    \includegraphics[width=0.5\linewidth]{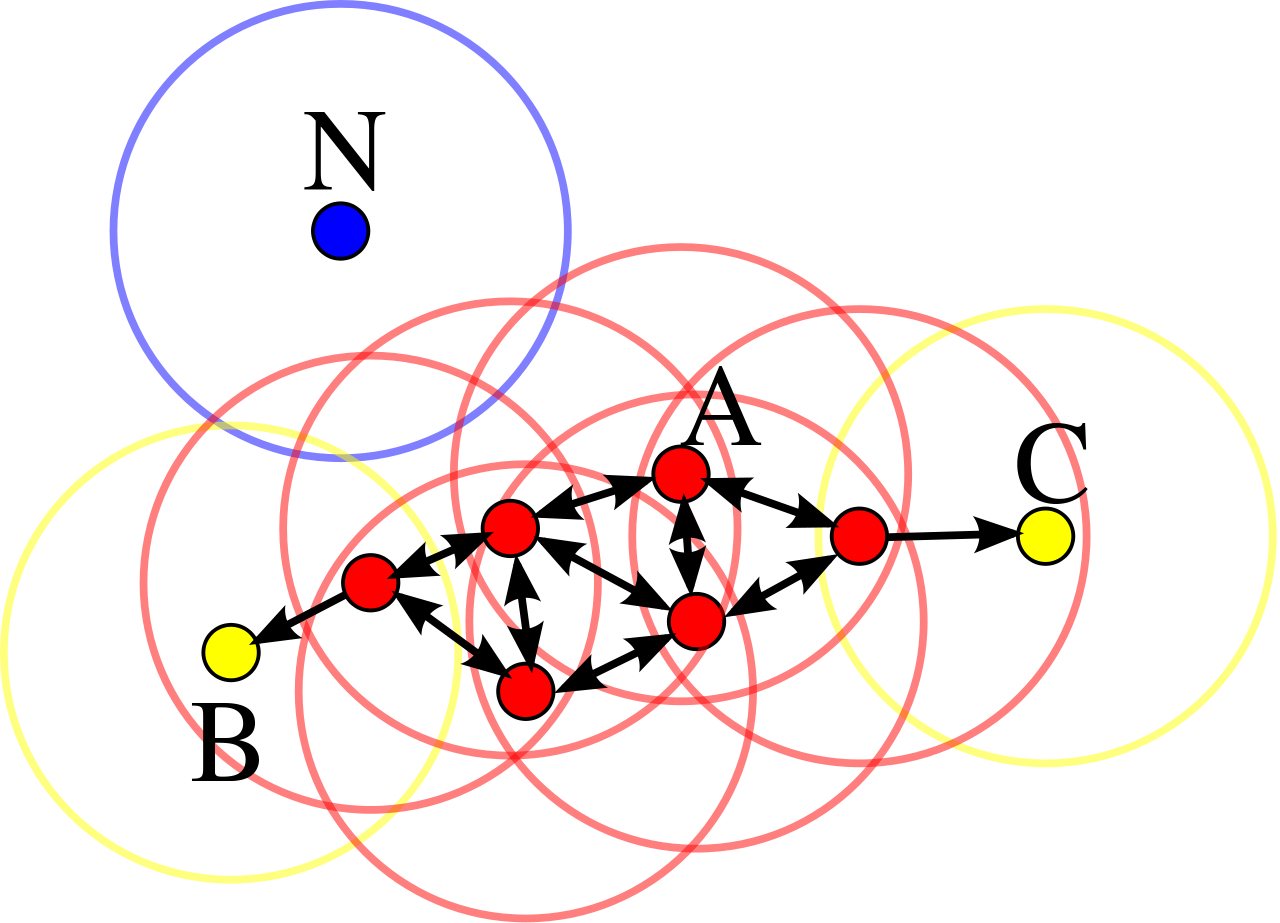}
    \caption{A circle is drawn around each point with radius $\varepsilon$. Based on a threshold (minPts, here 4) for the number of neighbors within the radius $\varepsilon$, a point is considered a core-point (e.g. A), edge-point (e.g. B, C), or an outlier (e.g. N). The direct density reachability is represented by the arrows. Observe how Points B and C are density connected, because both are density reachable from A. N is not density reachable, and thus considered to be a noise point \cite{Schubert2017DBSCANDBSCAN}}
    \label{fig:DBSCAN}
\end{figure}

To formally detect groups of rogue sensors in the learned representations, DBSCAN is applied on the learned representations. This is similar to \cite{Franceschi2019UnsupervisedSeries} who applied a classifier on the learned representation for classification purposes. While originally introduced in 1996 \cite{Ester1996ANoise}, DBSCAN is still a high-performing algorithm to this day that can cluster and detect outliers simultaneously \cite{Schubert2017DBSCANDBSCAN}. DBSCAN groups together points that are in a high-density region while it tags points that are located in a low-density region as outliers (see Figure \ref{fig:DBSCAN}). This fulfills our need of a method that can find clusters/groups of sensors in the latent space that show deviating behavior from the other sensors.

\subsection{Triplet network architecture: Dilated CNN}\label{sec:dilated_cnn}
\begin{figure}[h]
    \centering
    \begin{subfigure}{\linewidth}
    \centering
    \includegraphics[width=0.5\linewidth]{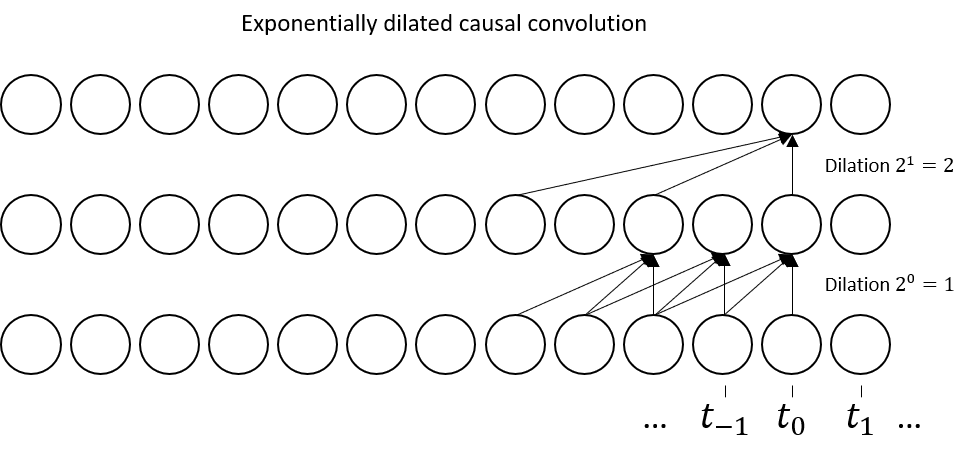}
    \caption{}
    \label{fig:dilated_cnn}
    \end{subfigure}
    \vfill
    \begin{subfigure}{\linewidth}
    \centering
    \includegraphics[width=0.5\linewidth]{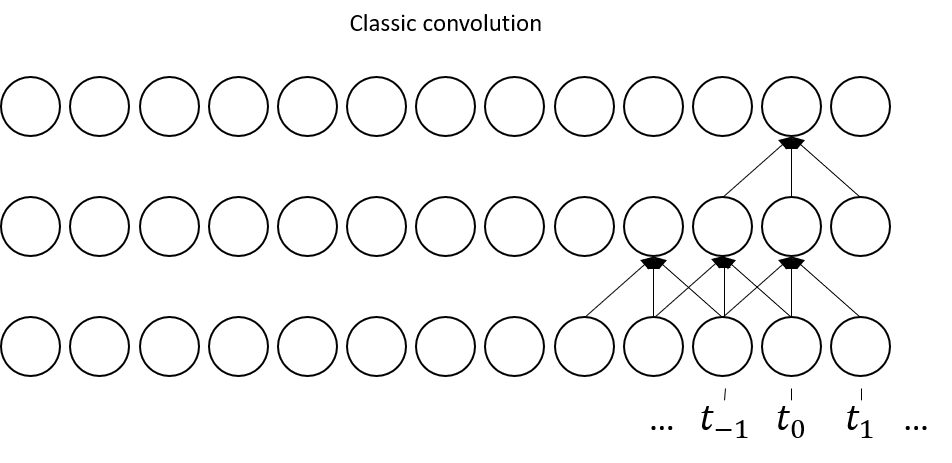}
    \caption{}
    \label{fig:classic_cnn}
    \end{subfigure}
    \caption{(\ref{fig:dilated_cnn}) shows the idea of an exponentially dilated causal convolution, repainted from \cite{Franceschi2019UnsupervisedSeries}. Observe how future values (e.g. $t_1$) in (\ref{fig:dilated_cnn}) are not included to calculate the output sequence. Moreover, thanks to the dilation long-range dependencies are easily captured in the output sequence. This is in contrast to the classic convolution in (\ref{fig:classic_cnn}) which uses future values to calculate the output sequence and does not include long-range dependencies.}
    \label{fig:cnn_overview}
\end{figure}

The triplet network is essentially characterized by the triplet loss (see section \ref{sec:triploss}) which means that the architecture of the network has no specific conceptual constraints (in contrast to other contrastive methods such as e.g. siamese networks which rely on a specific architecture). Given that IoT sensors produce temporal data, the only constraint is that the chosen architecture should be capable to process temporal data efficiently. Since the focus of our work is on the novel negative sampling technique introduced above, we rely on an existing proven encoder architecture as described in \cite{Franceschi2019UnsupervisedSeries,Bai2018AnModeling} based on exponential dilated convolutional neural networks. Dilated convolutional neural networks have proven to outperform traditional sequence modelling methods such as LSTMs in terms of scalability, the ability to capture (very) long-range dependencies, and more. The dilation distinguishes dilated CNNs from regular (1-dimensional) CNNs in that the latter takes into account future values during the convolutional operation while the former prevents the network from taking future values in the convolutional operations while including long-range temporal dependencies (see Figure \ref{fig:cnn_overview}). The output of the dilated CNN is then fed to a global max pooling layer and a fully connected layer to result in a vector of a fixed pre-determined size. For more details on the architecture, we refer to \cite{Franceschi2019UnsupervisedSeries}.

\section{Real world application}
We first introduce the data used to demonstrate our method in subsection \ref{sec:data}. Next, we describe how we trained and validated our method in subsection \ref{sec:training}. Lastly, in subsection \ref{sec:results} we discuss the results of our method applied on the data.
\subsection{Data}\label{sec:data}
We demonstrate our approach on a challenging dataset generated by Watermark soil moisture sensors (Irrometer Company, Inc., USA), retrieved from \cite{Janssens2011SensitivityClimate}. The soil moisture sensors were located on 3 different pear fields in Belgium and were operational from 2007-2009, measuring the soil water potential ($\Psi_{soil}$ in kPa) every 4 hours during the months April-September. $\Psi_{soil}$ is a negative value, expressed in pressure units (like kPa) that indicates how strongly the water is held in the soil pores by capillary and adhesive forces in a soil that is not saturated. Basically, the drier the soil, the stronger these forces are, and the more negative the soil water potential is, and the more difficult it is for plant roots to extract that water. Each field was divided into plots which each contained 5 sensors at different depth levels: 3 sensors at 30cm depth, 2 at 60cm depth, and 1 at 90cm depth. For training, we only use sensors of 30cm depth from 2 proximate fields (Bierbeek and Meensel) for the year 2007. This yields 63 sensors with each 884 observations. The reason for using a subset of sensors is that our method relies on the assumption that all sensors measure the same concept with the goal to find sensors that show deviating behavior w.r.t. other sensors. As such, these constraints ensure that variability in sensor measurements is not due to different depth levels, years, or location differences.
\subsection{Training and validation}\label{sec:training}
Following the guidelines from Franceschi et al. \cite{Franceschi2019UnsupervisedSeries}, we normalize the data by substracting the mean and dividing by the variance of the entire dataset. The number of negative samples $K$ is set to 6. A batch size of 3 (given the relatively small dataset) and 500 optimization steps translates to 24 epochs. We use 40 channels in all 10 intermediary layers of the network with a kernel size of 3. The causal network outputs a vector of length 60 which is further reduced to a vector in $\mathbb{R}^{2}$ through the fully-connected layer. The aforementioned parameters (except for the size of the output vector) are roughly similar to - and have been tested extensively in - the experiments in \cite{Franceschi2019UnsupervisedSeries}. The encoder is fit with the adam-optimizer and the corresponding default values as presented in \cite{Kingma2015Adam:Optimization}.

At last, the output in $\mathbb{R}^{2}$ is fed to the DBSCAN-algorithm. DBSCAN has two important parameters:
\begin{itemize}
    \item $\varepsilon$: the radius of the circle drawn around each point
    \item minPts: the minimum neighbors within the radius $\varepsilon$ of a point required, for that point to be considered a core point
\end{itemize}
It can easily be seen in Figure \ref{fig:DBSCAN} what the impact would be of different values for minPts and $\varepsilon$. To this end, we follow the heuristics outlined in \cite{Schubert2017DBSCANDBSCAN} to determine the values of minPts and $\varepsilon$. The value of minPts is set to 4 since the latent space is two-dimensional. For the value of $\varepsilon$, we construct the kneeplot as described in \cite{Ester1996ANoise,Schubert2017DBSCANDBSCAN}. Finally, the knee is located using the KneeLocator as described in \cite{Satopaa2011FindingBehavior}. The performance of our method is validated using expert knowledge - similar to~\cite{Mouret_Albughdadi_Duthoit_Kouam_Rieu_Tourneret_2021,Vilenski_Bak_Rosenblatt_2019} and formalized with the adjusted Rand index~\cite{Hubert_Arabie_1985}.
\subsection{Results}\label{sec:results}
\begin{figure}[h]
\centering
    \includegraphics[width=0.5\linewidth]{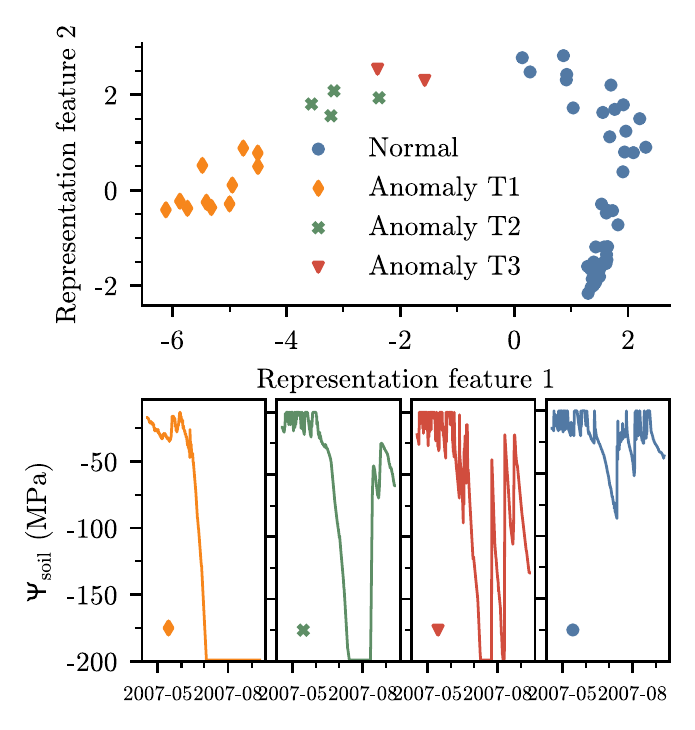}
  \caption{Learned representations in $\mathbb{R}^{2}$ with a corresponding sensor example for each type of anomaly as well as a normal sensor example. Orange: a sensor that hits the threshold and never recovers, green+red: sensors that hit the threshold and recover - respectively - slow and fast, blue: a normal sensor.}\label{fig2}
\end{figure}

After applying DBSCAN on the learned representations and through expert validation, three types of anomalies are detected in the space as shown in Figure \ref{fig2}: sensors that hit a sensor threshold and stay there (Anomaly T1), sensors that hit the threshold and "recover late" or "recover fast" (Anomaly T2 and T3 respectively), and sensors that are considered non-anomalous (Normal). Observe how type 2 (in green) and type 3 (in red) are close to each other. 
In fact, slightly adapting the parameters of DBSCAN would merge them into one cluster. The quality of the clusters containing the three types of sensor-measurements yielded by our method can be formalized with the adjusted Rand, comparing expert validation with our method's output. We report an adjusted rand index of 0.89. This indicates that the clusters generated by our method correspond well with the clusters that were expected based on the expert knowledge.
\section{Conclusion and future work}
In this paper, we present a fully self-supervised anomalous sensor detector. Our method builds on existing work for self-supervised representation learning of time-series using a contrastive loss, more specifically a triplet loss. We introduce a novel negative sampling technique based on Dynamic Time Warping-distance as to learn valid discriminative features in a setting where many sensors measure the same concept. The triplet loss is used in a proven dilated convolutional encoder architecture and deployed on sensor data from an agricultural context. Validation criteria were set up using domain expertise as to evaluate the performance of our method. After applying DBSCAN on the learned discriminative features, the different type of sensor anomalies present in the data, as defined by experts, are accurately separated from each other.

Based on the insights provided by our method, domain experts can decide to exclude or further investigate certain sensors. As such, our method can be used to reveal potential novel decision rules as to when a sensor should be considered rogue or not after our method highlights them to experts for further investigation. In future work, we will compare and benchmark our method to other recent methods. In addition, since the developed method is universal, it would also be interesting to demonstrate the use on other IoT datasets where multiple sensors measure a similar concept.

\bibliographystyle{splncs04}
\bibliography{references}
\end{document}